\documentclass[runningheads]{llncs}

\usepackage[utf8]{inputenc}
\usepackage{amsthm}
\usepackage{amsmath}
\usepackage{amssymb}
\usepackage{amsfonts}
\usepackage{easyeqn}
\usepackage[table]{xcolor}
\usepackage{algorithm}
\usepackage{algpseudocode}
\usepackage{listings}
\usepackage[multiple]{footmisc}
\usepackage{graphicx}
\usepackage{epstopdf}
\usepackage{bbold}
\usepackage{mathrsfs}
\usepackage{float}
\usepackage{indentfirst}
\usepackage{lscape}
\usepackage{booktabs}
\usepackage[flushleft]{threeparttable}

\makeatother
\def\data{D}
\def\bootData{\tilde{\data}}
\def\RR{\mathbb{R}}
\def\BB{\mathbb{B}}
\def\XX{\mathbb{X}}

\title{Geometry-Aware Maximum Likelihood Estimation of Intrinsic Dimension}
\author{Marina Gomtsyan\inst{1,2}
\and
Nikita Mokrov\inst{1,3} \and
Maxim Panov\inst{1}
\and 
Yury Yanovich\inst{1,4}
}
\authorrunning{M. Gomtsyan et al.}

\institute{Skolkovo Institute of Science and Technology, 121205 Moscow, Russia
\and
Higher School of Economics, 101000 Moscow, Russia 
\and
Moscow Institute of Physics and Technology, 141701 Moscow, Russia
\and
Institute for Information Transmission Problems, 127051 Moscow, Russia
\email{\{marina.gomtsyan, n.mokrov, m.panov, y.yanovich\}@skoltech.ru}\\
}

\begin{document}

\maketitle

\begin{abstract}
  
  
  The existing approaches to intrinsic dimension estimation usually are not reliable when the data are nonlinearly embedded in the high dimensional space. In this work, we show that the explicit accounting to geometric properties of unknown support leads to the polynomial correction to the standard maximum likelihood estimate of intrinsic dimension for flat manifolds. The proposed algorithm (GeoMLE) realizes the correction by regression of standard MLEs based on distances to nearest neighbors for different sizes of neighborhoods. Moreover, the proposed approach also efficiently handles the case of nonuniform sampling of the manifold. We perform numerous experiments on different synthetic and real-world datasets. The results show that our algorithm achieves state-of-the-art performance, while also being computationally efficient and robust to noise in the data.
  
  \keywords{Intrinsic dimension estimation \and Manifold learning \and Maximum likelihood estimation.} 
\end{abstract}

\section{Introduction}
\label{sec:intro}
  Dimensionality reduction is one of the critical steps of data analysis. The proper application of dimensionality reduction allows to decrease the required space for data storage and increase the speed of the data processing by machine learning algorithms. Most importantly, it often significantly improves the performance of many machine learning algorithms, which often rapidly degrades in high dimensions.
  
  The majority of existing dimensionality reduction methods require the true dimension of the data as an input parameter. Not surprisingly, the problem of estimating the true dimension of the data known as intrinsic dimension estimation is a well-studied problem, and numerous specialized intrinsic dimension estimation methods exist~\cite{Bailey1979,Grassberger1983,Levina2005,Hein2005,Lombardi2011,Little2012,Ceruti2014,Johnsson2015,Granata2016}. In addition, some dimensionality reduction methods such as principal component analysis (PCA)~\cite{Jolliffe1986} can be modified for estimating the intrinsic dimension, see~\cite{Fukunaga1971,Bishop1998,Tipping1999}. However, the existing intrinsic dimension estimation approaches have some disadvantages: some fail on data with a non-linear structure, some require a large number of observations for efficient performance, others are computationally expensive~\cite{Campadelli2015}.

  In this paper, we introduce a new efficient method for intrinsic dimension estimation. We base our approach on the \textit{Maximum likelihood estimation of intrinsic dimension} (MLE)~\cite{Levina2005} which is one of the most commonly used methods due to its simplicity and computational efficiency. However, when the true dimension of the data is large, the MLE method is known to underestimate it significantly. The explanation of this fact is contained in the key assumption of the method: the local neighborhood of each point is approximated by a linear subspace with a uniform density. Since real-world data often lies on or near to a nonlinear manifold with an arbitrary density, such an assumption is restrictive and leads to the bias in the procedure. To overcome the problems mentioned above we propose a data-driven approach, which explicitly introduces the correction for non-uniformity of density and nonlinearity of manifold into the likelihood and estimates unknown parameters by regression with respect to the radius of the neighborhood.

  Our main contributions are the following:
  \begin{itemize}
    \item We propose a new intrinsic dimension estimation method \textit{Geometry-aware maximum likelihood estimation of intrinsic dimension} (GeoMLE). Our approach takes into consideration the geometric properties of a manifold and corrects for a nonuniform sampling.

    \item GeoMLE shows the state-of-the-art results in the estimation of intrinsic dimension. In numerous experiments, GeoMLE outperforms MLE~\cite{Levina2005} and other intrinsic dimension estimators. In particular, our estimator gives accurate results for datasets in high dimensions, in case of which the performance of many competitors is rather weak. The following link provides access to the implementation of the proposed method and all the experiments:~\url{https://github.com/premolab/GeoMLE}.
  \end{itemize}
  
\section{Maximum Likelihood Estimator of Intrinsic Dimension}
\label{sec:mle}
  Consider data manifold of unknown dimension \(m\):
  \begin{EQA}[c]
    \XX = \{x = g(b) \in \RR^p\colon b \in \BB \subset \RR^m\},
  \end{EQA} 
  where \((\BB, g)\) is a single coordinate chart embedded into an ambient \(p\)-dimension space \(\RR^p\), such that \(m \leq p\). The mapping \(g\) is a one-to-one mapping from an open bounded set \(\BB \subset \RR^p\) to manifold \(\XX = g(\BB)\), with a differentiable inverse map \(g^{-1}\colon \XX \rightarrow \BB\). The manifold \(\XX\) is unknown, and a finite data set \(\data = \{X_1, \dots, X_n\} \subset \XX \subset \RR^p\) is sampled from a distribution with an unknown density \(f(x)\). We note that the single coordinate chart is a technical simplification, and the results are correct at least for manifolds covered with finite atlases.
  
  Levina and Bickel~\cite{Levina2005} suggested to consider the binomial process
  \begin{EQA}[c]
    N(t, x) = \sum_{i = 1}^n \mathbb{1}\{X_i \in S_x(t)\}, ~~ 0 \leq t \leq R,
  \end{EQA}
  where \(S_x(t)\) is a ball of radius \(t\) centered at \(x\). They approximate propose to this process by Poisson process \(N_{\lambda}(t, x)\) with rate \(\lambda_{m, \theta}(t)\) and \(\theta = \log f(x)\). Suppressing the dependence on \(x\), the log-likelihood of the observed process \(N_{\lambda}(t, x)\) is 
  \begin{EQA}[c]
    L_{\lambda}(m, \theta) = \int_0^R \log \lambda_{m, \theta}(t) d N(t) - \int_0^R \lambda_{m, \theta}(t) dt.
  \label{poisson_likelihood}
  \end{EQA}
  The key idea of MLE~\cite{Levina2005} is to fix a point \(x\) and for an unknown smooth density \(f\) on \(\XX\) assume that \(f(z) \approx \text{const}\) in a ball \(z \in S_x(R) \subset  \RR^p\) of small radius \(R\), while the intersection  of \(\XX\) and \(S_x(R)\) is approximated by \(m\)-dimensional ball \(S_x^m(R)\). Then, the observations are treated as a Poisson process in \(S_x^m(R) \subset \RR^m\).
  The rate of the Poisson process for the resulting approximation is
  \begin{EQA}[c]
    \hat{\lambda}_{m, \theta}(t) = f(x) V_m m t^{m - 1},
  \label{poisson_rate_const}
  \end{EQA}
  where \(V_m\) is the volume of the unit sphere in \(\RR^m\).

  Let \(T_k(x)\) be the Euclidean distance from a fixed  point \(x\) to its \(k\)-th nearest neighbor in the sample \(\data\). We state the following Proposition~\cite{Levina2005}.

  \begin{proposition}
  \label{prop:mle}
    The intrinsic dimension estimate for a manifold \(\XX\) at a point \(x\) obtained by maximizing the likelihood~\eqref{poisson_likelihood} with a rate~\eqref{poisson_rate_const} is equal to 
    \begin{EQA}[c]
      \hat{m}_{R}(x) = \Bigg(\frac{1}{N(R, x)} \sum_{j = 1}^{N(R, x)} \log \frac{R}{T_j(x)}\Bigg)^{-1}.
    \end{EQA}
  \end{proposition}
  The proof of the proposition can be found in supplementary materials.
  For numerical calculations it might be more convenient to fix the number of neighbors \(k\) rather than the radius of the ball \(R\). Then the MLE reads as
  \begin{EQA}[c]
    \hat{m}_{k}(x) = \Bigg(\frac{1}{k - 1} \sum_{j = 1}^{k - 1} \log \frac{T_k(x)}{T_j(x)}\Bigg)^{-1},
  \end{EQA}
  where \(k\) is the number of neighbors. 

\section{Geometry-Aware MLE of Intrinsic Dimension}
  Levina and Bickel~\cite{Levina2005} approximate the local neighborhood of each point by a linear subspace with a uniform density. However, usually, real-world data lies on or near to an unknown nonlinear manifold with a density far from being uniform, which leads to bias in the MLE method. In this section, we propose an improvement of the MLE by introducing a correction for non-uniformity of density and nonlinearity of manifold into the likelihood function.
  
\subsection{Adjusted Likelihood Construction}
  We start from the general Poisson process-based likelihood~(\ref{poisson_likelihood}) but aim to find a better approximation to the rate \(\lambda_{m, \theta}(t)\). Our derivation requires several assumptions of manifold \(\XX\) and density \(f(x)\).

  We assume that density \(f(x)\) is bounded for \(x \in \XX\) and denote \(f_{\max} = \sup\limits_{x \in \XX} f(x)\). Let us also define the bounds on maximum eigenvalues of first and second derivatives of \(f(x)\):
  \begin{EQA}[c]
    C_{p, 1} = \sup\limits_{x \in \XX, \theta \in T_x(\XX)\colon \|\theta\| = 1} \|\nabla_{\theta} f(x)\|, \qquad
    C_{p, 2} = \sup\limits_{x \in \XX, \theta \in T_x(\XX)\colon \|\theta\| = 1} \|\nabla_{\theta} \nabla_{\theta} f(x)\|,
  \end{EQA}
  where \(T_x(\XX)\) is a tangent space to the manifold \(\XX\) at the point \(x \in \XX\). 
  
  We also assume that the manifold \(\XX\) is not too curved. This limitation can be expressed in terms of the second normal form \(\mathrm {I\!I} (\theta, \theta)\) and the Ricci curvature \(\operatorname{Ric}(\theta, \theta)\), those are bounded for manifolds with smooth enough parametrizations according to Lemmas 3 and 4 from~\cite{Yanovich2016}. We assume that for a given manifold \(\XX\) there exist such positive constants \(C_{ \mathrm {I\!I}}\) and \(C_{\operatorname{Ric}}\) that for all \(x \in \XX\), \(\theta \in T_x(\XX)\), and \(\|\theta\| = 1\) it holds
  \begin{EQA}[c]
    \mathrm {I\!I} (\theta, \theta) \leq C_{\mathrm {I\!I} }, \quad \operatorname{Ric}(\theta, \theta) \leq C_{\operatorname{Ric}}.
  \end{EQA}
  
  %

  \begin{proposition}
  \label{prop:rate}
    The rate of Poisson process \(N_{\lambda}(t, x)\) on the manifold \(\XX\) can be expressed as 
    \begin{EQA}[c]
      \lambda_{m, \theta}(R) = R^{m - 1} V_m (m f(x) + R^2 \delta(R)) = \hat{\lambda}_{m, \theta}(R) + R^{m + 1}V_m \cdot \delta(R), 
    \end{EQA}
    where the term \(\delta(R)\) can be bounded as
    \begin{EQA}[c]
      |\delta(R)| \leq 8 f_{\max} (m + 2) \frac{m C_{\mathrm {I\!I}}}{24} + C_{p, 2}(m + 2) + (m + 3)R C_{p, 1}C_{\operatorname{Ric}} \\
      + (m + 4) R^2 C_{p, 2} C_{\operatorname{Ric}} + f(x) C_{\operatorname{Ric}} (m + 2).
    \label{mle_bias}
    \end{EQA}
  \end{proposition}
  The result of Proposition~\ref{prop:rate} allows us to lower bound the true log-likelihood~\eqref{poisson_likelihood} by the following function:
  \begin{EQA}[c]
    \hat{L}(m, \theta) = (m - 1) \int_0^R \log t ~ d N(t, x) + N(R, x) \log V_m + N(R, x) \log m \\ 
    + N(R, x) \log f(x) + \int_0^R \log(2t^2 \delta) d N(t, x)
    - V_m R^m \left(f(x) + \frac{R^2 \delta(R)}{m + 2}\right).
  \end{EQA}
  The following result allows to compute the maximizer for the function \(\hat{L}(m, \theta)\).
  \begin{proposition}
  \label{prop:geo_mle}
    The maximum of the function \(\hat{L}(m, \theta)\) is achieved by
    \begin{EQA}[c]
      \breve{m}_R(x) = \hat{m}_R(x) \left(1 + \delta(R) \frac{R^2}{N(R, x)}\right).
    \label{geo_mle}
    \end{EQA}
  \end{proposition}
  Unfortunately, the estimate \(\breve{m}_R(x)\) cannot be computed directly as the quantity \(\delta(R)\) is unknown. We also know the explicit upper bound~\eqref{mle_bias} on \(\delta(R)\), but it still includes a number of unknown parameters depending on manifold \(\XX\) and density \(f(x)\).

  However, the form of dependency in equation~\eqref{geo_mle} suggests that we can try to find \(\breve{m}_R(x)\) by computing the correction to the standard MLE \(\hat{m}_R(x)\). We note that by Taylor expansion we can represent~\eqref{geo_mle} in the following form
  %
  %
  \begin{EQA}[c]
    \breve{m}_R(x) = \hat{m}_R(x) + P_{l, \eta}(R) + O(R^{l + 1}),
  \label{geo_mle_poly}
  \end{EQA}
  where \(P_{l, \eta}(R)\) is a polynomial of degree \(l\) with the constant term equal to zero and other coefficients given by vector \(\eta\).

  The key idea is to consider the estimates \(\hat{m}_R(x)\) for different values of \(R\) and try to fit polynomial approximation to them. Under the assumption that \(\breve{m}_R(x)\) does not depend on \(R\), the zero order term in the approximation will give an estimate \(\breve{m}(x)\) of the intrinsic dimension. By fixing the number of neighbors \(k\) and estimating \(\hat{m}_k(x)\) we obtain the following polynomial regression problem
  \begin{EQA}[c]
    \hat{m}_k(x) = \breve{m}(x) + P_{l, \eta}(T_k(x)) + \epsilon_k,
  \label{geo_mle_problem}
  \end{EQA}
  where \(\epsilon_k\) represents an error due to ignoring higher order terms in polynomial approximation. The estimation of \(\breve{m}(x)\) and coefficients of polynomial \(P_l\) can be done based on estimates \(\hat{m}_k(x)\) computed for different values of the number of neighbors \(k\) and corresponding distances \(T_k(x)\). 
    
\subsection{Algorithmic implementation of GeoMLE}
\label{sec:algo}
  To estimate the intrinsic dimension \(\breve{m}(x)\) of the manifold in the vicinity point \(x\) based on the sample \(\data = \{X_1, \dots, X_n\}\) by polynomial regression, we should construct a dataset of MLEs \(\hat{m}_{k_1}(x), \dots, \hat{m}_{k_2}(x)\) for a range of values of \(k = k_1 \leq \dots \leq k_2\) with \(k_1\) and \(k_2\) being input parameters of the method. It is important to choose \(k_1\) large enough to ensure the stability of distance estimates \(T_k(x)\), while \(k_2\) can not be very large to validate the approximations used to construct the estimates. In practice, due to the finite size of the data, the estimates \(\hat{m}_{k}(x)\) are unstable for small and even moderate values of \(k\). We suggest to estimate this uncertainty by special bootstrap procedure and incorporate obtained uncertainty estimates directly into regression problem. Such an approach also allows making the method less dependent on the choice of the number of nearest neighbors \(k\).

  We start by creating \(M\) bootstrapped datasets \(\bootData_1, \dots, \bootData_M\) of the sample \(\data = \{X_1, \dots, X_n\}\). For each \(k\) we repeat the following procedure. First, we find \(k\) nearest neighbors of point \(x\) among the points in \(\bootData_j\) bootstrapped dataset for \(j = 1, \dots, M\). Then, for \(x\) we calculate its distance from its \(k\)-th nearest neighbor \(T_k(x, \bootData_j)\) in \(\bootData_j\) and find its dimension \(\hat{m}_k (x, \bootData_j)\) by MLE approach.
    
\begin{figure}[t!]
    \centering
    \includegraphics[width=.8\textwidth]{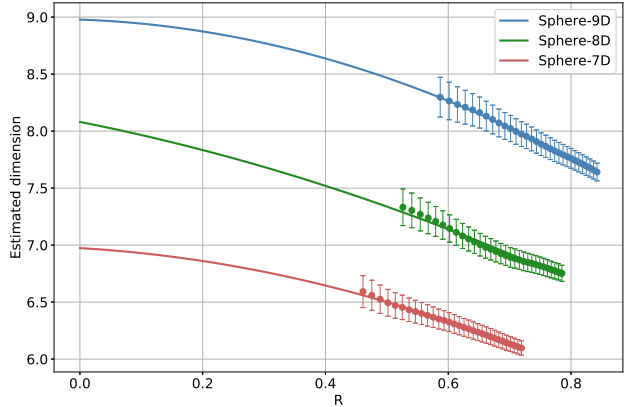}
    \caption{Illustration of GeoMLE for the samples from spheres of 3 different dimensions. Different colors of points indicate average MLEs of bootstraped datasets for corresponding \(R\) with corresponding standard deviations. Curves show corresponding quadratic regression fitted to the points.}
  \label{fig:GeoMLE}
  \end{figure}

  After that, we average distances to neighbors and MLEs in the following way 
  \begin{EQA}[c]
    \bar{T}_k(x) = \frac{1}{M}\sum_{j = 1}^M T_k(x, \bootData_j), ~~ 
    \bar{m}_k(x) = \frac{1}{M}\sum_{j = 1}^M \hat{m}_k(x, \bootData_j).
  \end{EQA}
  In addition, for each neighbor \(k\) we calculate variances of MLE dimensions for \(x\) in the sample
  \begin{EQA}[c]
    \hat{\sigma}_k^2(x) = \frac{1}{M} \sum_{j = 1}^{M} \bigl(\hat{m}_k(x, \bootData_j) - \bar{m}_k(x)\bigr)^2.
  \end{EQA} 
  Given estimates of variances \(\hat{\sigma}_k^2(x)\) of estimated dimension \(\hat{m}_k (x)\), we can build a heteroscedastic polynomial regression model
  \begin{EQA}[c]
    \min_{\breve{m}(x), \eta} \sum_{k = k_1}^{k_2} \frac{1}{\hat{\sigma}_k^2(x)} \Big(\bar{m}_k(x) - \breve{m}(x) - P_{l, \eta}\bigl(\bar{T}_k(x)\bigr)\Big)^2,
  \end{EQA}
  where \(P_{l, \eta}\) is the polynomial of degree \(l\) with constant term equal to zero and other coefficients given by vector \(\eta\). In order to find the resulting intrinsic dimension \(\breve{m}\) we can run the procedure for each point in the sample \(\data = \{X_1, \dots, X_n\}\) and average the obtained local estimates:
  \begin{EQA}[c]
    \breve{m} = \frac{1}{n} \sum_{i = 1}^n \breve{m}(X_i).
  \end{EQA}
  Figure~\ref{fig:GeoMLE} illustrates GeoMLE approach by showing resulting polynomial estimates for the samples from spheres of three different dimensions.

\section{Experiments}
  In this section, we present the performance of GeoMLE by conducting the series of experiments on synthetic and real-world datasets that are suggested as a benchmark for evaluating intrinsic dimension estimators in~\cite{Rozza2012}. Simulated datasets used in our experiments are generated from different well-known manifolds such as linear subspace with normal distribution, sphere, Swiss roll, helix, cube surface, paraboloid, and some others. For the experiments on synthetic data we take the size of datasets equal to 1000. Real-world datasets in our experiments include Digits~\cite{Kaynak1995}, ISOMAP face~\cite{Tenenbaum2000}, and ISOLET~\cite{Fanty1990}. 
  In our experiments we consider several classical baseline methods such as Local PCA~\cite{Fukunaga1971}, \(\text{MiND}_{\text{KL}}\)~\cite{Lombardi2011} and MLE~\cite{Levina2005}, and state-of-the-art approaches DANCo~\cite{Ceruti2014} and ESS~\cite{Johnsson2015} according to the recent review~\cite{Campadelli2015}. See a more detailed discussion of these methods in Section~\ref{sec:related_work}. The quadratic polynomial was used for the solution of GeoMLE.

\subsection{Simulated and real-world data}  
  Table~\ref{sim_data} presents the resulting estimates for real-world and selected synthetic datasets. Here \(p\) denotes the full dimension of data space and \(m\) is the true dimension of the data for synthetic datasets, while for the real-world datasets \(m\) denotes the dimension determined by experts since true dimensions for real-world datasets are not known. The results are averaged over 10 independent samples, and best estimates for each dataset are in bold. It is clearly seen that GeoMLE is the most accurate estimate in the majority of cases, while other methods give the best results only for few datasets each. 

  \begin{table}[t]
    \centering
    \caption{Estimation results achieved on synthetic and real-world datasets. \(p\) is the dimension of space into which the data is embedded and \(m\) is the true dimension of the data.}
    \begin{tabular}{lcccccccc}
      \toprule
      Dataset & \(p\) & \(m\) & MLE & GeoMLE  & \(\text{MiND}_{\text{KL}}\) & DANCo & ESS & PCA  \\
      \midrule
      Affine  & 10 & 10 & 8.0  & \textbf{10.0} & 8.0 & 9.8  & 10.2 & \textbf{10.0}  \\
      Cubic & 35 & 30 & 19.8 & \textbf{29.8} & 20.4 & 30.8 & 31.2 & 31.0 \\
      Helix & 3 & 1 & \textbf{1.0} & \textbf{1.0} & \textbf{1.0} & \textbf{1.0} & 3.0 & 3.0 \\
      Helix & 13 & 2 & 3.0 & \textbf{2.4} & 3.0 & 3.0 & 2.8 &  3.0 \\
      Moebius & 3 & 2 & \textbf{2.0} & \textbf{2.0} & \textbf{2.0} & \textbf{2.0} & \textbf{2.0} & 3.0 \\
      Nonliner & 36 & 6 & 7.0 & 6.6 & \textbf{6.2} & 8.0 & 12.0 & 12.0 \\
      Norm & 50 & 50 & 27.0 & \textbf{50.0} & 28.8 & 47.0 & 50.2 & \textbf{50.0} \\
      Paraboloid  & 30  & 9 & 6.0 & \textbf{9.0} & 6.0 & 8.0  & 1.0 & 1.0 \\
      Roll & 3 & 2  & \textbf{2.0} & 2.6 & \textbf{2.0} & \textbf{2.0} & 3.0 & 3.0 \\
      Sphere & 15 & 10 & 9.0 & \textbf{9.8} & 9.0 & 11.4 & 11.0 & 11.0 \\
      Spiral & 3 & 1  & 2.0  & 1.2  & \textbf{1.0} & \textbf{1.0} & 2.0 & 2.0 \\
      Uniform & 55 & 50 & 27.0 & 49.8 & 28.6 & 51.2 & 49.4 & \textbf{50.0} \\
      \midrule
      Isomap & 4096 & 3  & 4  & \textbf{3.3} & 4.0 & 6.0 & 7.4 & 10.0 \\
      Digits & 64 & 9-11 & 7.7 & \textbf{11.0}  & 8.0 & \textbf{9.0} & 13.2 & 23.0 \\
      ISOLET & 617 & 16-22 & \textbf{16.9} & 25.0 & 15.0 & 14.0 & 12.4 & 13.0 \\
      \bottomrule
    \end{tabular}
    \label{sim_data}
  \end{table}
  


  \begin{figure}[t]
    \centering
    \includegraphics[width=.8\textwidth]{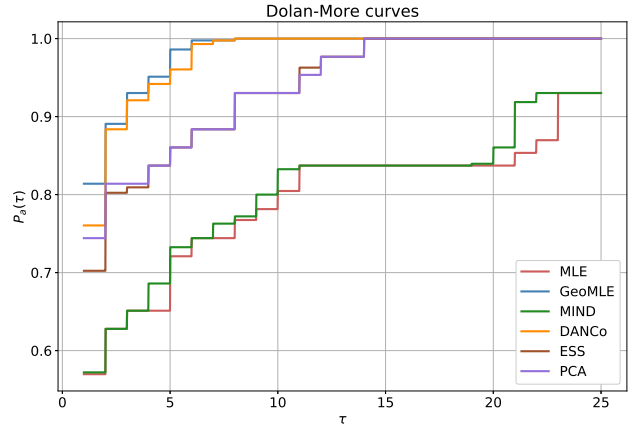}
    \caption{Dolan-More curves for all synthetic datasets to compare the estimates of MLE, GeoMLE, \(\text{MiND}_{\text{KL}}\), DANCo, ESS, and PCA. \(p_a(\tau)\) shows the ratio of problems on which the performance of the \(a\)-th method is the best.}
  \label{fig:figure2}
  \end{figure}
  
  In Figure~\ref{fig:figure2} we summarize the results for synthetic datasets by plotting Dolan-More curves~\cite{Dolan2002} which are a benchmarking tool for comparison of the performance of different methods. Each curve \(p_a(\tau)\) defines the fraction of problems in which the \(a\)-th algorithm has the error not more than \(\tau\) times bigger than the best competitor. Thus, the higher curve, the better performance of the algorithm, and \(p_a(1)\) is equal to the fraction of problems for which algorithm \(a\) gives the best result over all the algorithm. For evaluation we consider 45 different synthetic datasets with 10 independent samples generated for each of them. We see that GeoMLE shows the best result in more than 80\% of the problems. The closest competitor to GeoMLE is DANCo, while other methods perform significantly worse.
  


\subsection{Robustness to noise}
\begin{figure}[tb]
    \centering
    \includegraphics[width=.8\textwidth]{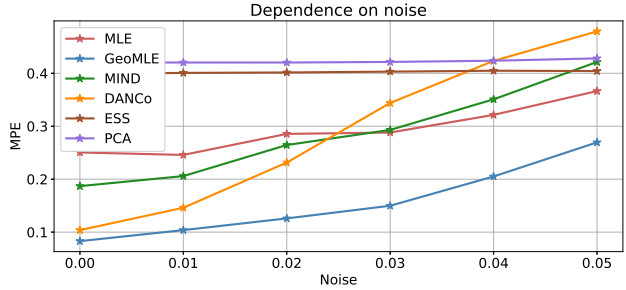}
    \caption{Dependence of estimates of MLE, GeoMLE, \(\text{MiND}_{\text{KL}}\), DANCo, ESS, and PCA on noisy  4-dimensional sphere data.}
    \label{fig:figure3}
  \end{figure}

  We also evaluate the robustness of GeoMLE and other methods with respect to noise. We add zero mean Gaussian noise to samples for synthetic datasets. Standard deviations of noise are taken to be from \(0\) to \(0.05\) with step size equal to \(0.01\). For evaluation of the results we calculate mean percentage error (MPE), which is \(\text{MPE} = \frac{1}{n} \sum_{i = 1}^n \frac{|m_i - \hat{m}_i|}{m_i}\), where \(n\) is the number of synthetic manifolds, \(m_i\) is the true dimension, and \(\hat{m}_i\) is the estimated dimension. The results are averaged over all synthetic datasets and 5 independent realizations of noise. We see in Figure~\ref{fig:figure3} that PCA and ESS are almost not affected by noise, while GeoMLE still shows the best quality of intrinsic dimension estimation for considered levels of noise. Interestingly, DANCo's performance decreases most rapidly with increased noise level. 

\subsection{Effect of nonuniform sampling}
  Finally, we want to explicitly test whether GeoMLE allows to correct for nonuniform density, as in all the previous synthetic experiments density was always uniform. In Table~\ref{densities} we compare the performance of GeoMLE and MLE on 5-dimensional spheres with uniform and nonuniform densities embedded into 7-dimensional space. Non-uniformity was achieved by generating points with uniform density in 5 dimensional space and then projecting them on the sphere. The presented estimates are averaged over 10 samples of 1000 points each. Despite there are no major differences between the methods for spheres with uniform densities, in case of nonuniform densities MLE underestimates the dimension while GeoMLE gives much more accurate result. 
    
  \begin{table}[t]
    \centering
    \caption{Dimension estimates of GeoMLE and MLE of 5-dimensional sphere in 7-dimensional space with uniform and nonuniform densities. The results are averaged over 10 samples of 1000 points each.}
    \begin{tabular}{lcc}
      \toprule
      Method &  Uniform & Nonuniform \\
      \midrule
      GeoMLE & 5.1 & 4.9 \\
      MLE & 4.8 & 4.6 \\
      \bottomrule
    \end{tabular}
    \label{densities}
  \end{table}

\section{Related Work}
\label{sec:related_work}
  This section reviews most recent and efficient intrinsic dimension estimators, which can be classified into 4 big groups: projective, fractal, nearest neighbor based, and simplex based.
  
  Projective intrinsic dimension estimation methods are based on Multidimensional Scaling (MDS)~\cite{Romney1972} that try to maintain as much as possible pairwise distances in the data, and Principal Component Analysis (PCA)~\cite{Jolliffe1986}, that find the best projection subspace. One of the most efficient methods in this group is local PCA~\cite{Fukunaga1971}.

  Fractal methods rely on the assumption that data points are sampled through some smooth probability density function from an underlying manifold. Two of the widely used fractal methods are Correlation dimension~\cite{Grassberger1983} and the method by Hein and Audibert~\cite{Hein2005}.

  The main assumption of nearest neighbor based approaches is that close points are uniformly drawn from \(m\)-dimensional balls with sufficiently small radii, where \(m\) is the true dimension of the data. Some of the most successful nearest neighbor based methods are MLE~\cite{Levina2005}, \(\text{MiND}_{\text{KL}}\)~\cite{Lombardi2011}, and DANCo~\cite{Ceruti2014}.  \(\text{MiND}_{\text{KL}}\)~\cite{Lombardi2011} computes the empirical probability density function of the neighborhood distances. Then, it finds the distribution of the neighborhood distances computed from points uniformly drawn from synthetic hyperspheres of known dimension. The idea of \(\text{MiND}_{\text{KL}}\) is to minimize the Kullback-Leibler divergence between these two distributions to obtain the dimension estimate. DANCo~\cite{Ceruti2014} is an extension of \(\text{MiND}_{\text{KL}}\) and reduces the underestimation, which is the main downside of \(\text{MiND}_{\text{KL}}\). Besides the probability density function modeling the distribution of nearest neighbor distances, DANCo adds a second probability density function modeling the distribution of pairwise angles.
  
  Finally, simplex based methods evaluate simplex volumes and then analyze their geometric properties. One of the best performing methods in this category is Expected Simplex Skewness (ESS)~\cite{Johnsson2015}.

\section{Conclusions}
\label{sec:conclusions}
  In this paper we have introduced a state-of-the-art intrinsic dimension estimator GeoMLE. It was inspired by one of the most widely used intrinsic dimension estimation approaches suggested by Levina and Bickel~\cite{Levina2005}. We extended the method by taking into consideration geometric properties of unknown support and possible non-uniformity of the data sampling. In the result, we propose a data-driven correction which allows to overcome the main drawbacks, which are underestimation of the true dimension in high dimensions and sensitivity to nonuniform sampling. 
  
  We compare the performance of GeoMLE to other intrinsic dimension estimators in the variety of synthetic and real-world problems. The comparison shows that GeoMLE achieves state-of-the-art performance with DANCo~\cite{Ceruti2014} being its closest competitor. Moreover, our approach is computationally faster than DANCo, while also being more robust to noise. 
  
\bibliographystyle{splncs04}
\bibliography{bibliography}

\appendix

\section{Proof of Proposition~\ref{prop:mle}}
\label{sec:proof_mle}
  Let us consider the inhomogeneous binomial process \(\{N(t, x), 0 \leq t \leq R\}\),
  \begin{EQA}[c]
    N(t,x) = \sum_{i=1}^n \mathbb{1} \{X_i \in S_x(t)\},
  \end{EQA}
  that counts observations within distance \(t\) from \(x\). Let \(T_k(x)\) be the Euclidean distance from a fixed  point \(x\) to its \(k\)-th nearest neighbor in the sample. This process can be approximated by a Poisson process. The rate \(\lambda(t)\) of the process \(N(t)\) can be written as
  \begin{EQA}[c]
    \lambda(t) = \frac{\partial P(x \in S_x(r))}{\partial r} |_{ r = t}.
  \end{EQA}
  Since the density \(f(x)\) in \(S_x(t)\) is approximated by a constant and \(V(m) = \pi^{m / 2}(\Gamma(m/2 + 1))^{-1}\), which is the volume of a unit sphere in \(\RR^m\), it follows that 
  \begin{EQA}[c]
    \lambda(t) \approx \hat{\lambda}(t) = f(x) V(m) m t^{m - 1},
    \label{eqn:lambda}
  \end{EQA}
  since \(\frac{d (V(m) t^m)}{d t}= V(m) m t^{m - 1}\) is the surface area of the sphere \(S_x(t)\). Letting \(\theta = \log f(x)\), we state the following Proposition. The log-likelihood of the observed process \(N(t)\) can be written as 
  \begin{EQA}[c]
    L(m, \theta) = \int_{0}^{R} \log \hat{\lambda}(t) d N(t) - \int_{0}^{R} \hat{\lambda}(t) d t \\
    = (m - 1) \int_{0}^{R} \log t d N(t) + N(R) \log V(m) \\ 
    + N(R) \log(m f(x)) - V(m) R^m f(x).
  \end{EQA}
  MLEs must satisfy the equation
  \begin{EQA}[c]
    \frac{\partial L}{\partial \theta} = N(R) - e^{\theta} V(m) R^m = 0,
  \end{EQA}
  from which it is obtained that \(e^{\theta} = \frac{N(R)}{V(m) R^m}\), and the equation 
  \begin{EQA}[c]
    \frac{\partial L}{\partial m} = \int_{0}^{R} \log t d N(t) + N(R) \frac{V'(m)}{V(m)} + \frac{N(R)}{m} \\
    - V'(m) R^m \frac{N(R)}{V(m) R^m} - V(m) R^m \frac{N(R)}{V(m) R^m} \log R \\
    = \int_{0}^R \log t d N(t) + \frac{N(R)}{m} - N(R) \log R = 0,
  \end{EQA}
  where \(N(R) = N(R,x)\). Thus 
  \begin{EQA}[c]
    \hat{m}_{R}(x) = \frac{1}{\log R - \frac{1}{N(R, x)} \int_{0}^R \log t d N(t)}
    = \Bigg(\frac{1}{N(R, x)} \sum_{j = 1}^{N(R, x)} \log \frac{R}{T_j(x)}\Bigg)^{-1}.
  \end{EQA}

\section{Proof of Proposition~\ref{prop:rate}}
\label{sec:proof_geo_mle}
  The manifold \(\XX\) is generally nonlinear and density \(f(x)\) is non-constant.
  Let us estimate  \(\frac{\partial P(x \in S_x(R))}{\partial R}\mid_{R = t}\) by considering the results obtained in~\cite{Yanovich2016,Yanovich2017}. Firstly, we replace the domain of integration with sphere \(\tilde{S}_{\tilde{X}}(R)\) in tangent space  \(T_X(\XX)\) and calculate the error of this replacement. From Lemma~8~\cite{Yanovich2016} we know that
  \begin{EQA}
    & & \left| P(x \in S_{\tilde{X}}(R + \Delta R)) - P(x \in \tilde{S}_{\tilde{X}}(R + \Delta R) - P(x \in S_{\tilde{X}}(R)) + P(x \in \tilde{S}_{\tilde{X}}(R)) \right| \\
    & \leq & 8 V_m f_{max} \left( (R + \Delta R)^{m + 2} - R^{m + 2} \right) \frac{m C_{II}}{24} \\ 
    & = & 8 V_m f_{max} \Delta R \frac{m C_{II}}{24}\sum_{i = 0}^{m + 1}((R + \Delta R)^i r^{1 - i - 1}) \\
    & \leq & 8 V_m f_{max}(m + 2) \Delta R (R + \Delta R)^{m + 1} \frac{m C_{II}}{24}.  
  \end{EQA}
  We replace the density \(f(\tilde{X})\) with the density at a point \(f(x)\) and calculate the error of this replacement 
  \begin{EQA}
    & & \left| \int_{\tilde{S}_{\tilde{X}}(x)} f(R) d V(\tilde{X}) -  \int_{\tilde{S}_{X}(R)} f(x) d V(\tilde{X})\right| \\
    & = & \left\{f(x) = p(x) + t \nabla_{\theta} p(x) + t^2 / 2 \nabla_{\tilde{\theta}} \nabla_{\tilde{\theta}} p(\tilde{\tilde{X}}), \tilde{\tilde{X}} \in \tilde{S}_{X}(R) \right\} \\
    & = & \int_{S^{q - 1}} \int_0^r (t\nabla_{\theta} p(x) + t^2 / 2 \nabla_{\tilde{\theta}} \nabla_{\tilde{\theta}} p(\tilde{\tilde{X}})) (t^{q - 1} + t^{q + 1} Ric_{\tilde{\tilde{X}}} (\tilde{\theta}, \tilde{\theta}))d t s \theta \\
    & \leq & \int_{A^{m - 1}} \int_0^R t^m \nabla_{\theta} p(x) d t d \theta + \int_{A^{m - 1}} \int_0^R t^{m + 1} / 2 \nabla_{\tilde{\theta}} \nabla_{\tilde{\theta}} p(\tilde{\tilde{X}}) d t d \theta \\ 
    & + & \int_{A^{m - 1}}\int_0^R t^{m + 2}\nabla_{\theta} p(x) Ric_{\tilde{\tilde{X}}} (\tilde{\theta}, \tilde{\theta}) d t d\theta \\
    & + & \int_{A^{m - 1}} \int_0^R t^{m + 3} / 2 \nabla_{\tilde{\theta}} \nabla_{\tilde{\theta}} p(\tilde{\tilde{X}}) Ric_{\tilde{\tilde{X}}} (\tilde{\theta}, \tilde{\theta}) d t d \theta \\ 
    & \leq & R^{m + 2} V_m (C_{p, 2} + R C_{p, 1}C_{Ric} + R^2 C_{p, 2} C_{Ric}).
  \end{EQA}
  We further bound
  \begin{EQA}
    & & \left| \int_{\tilde{S}_{X} (R + \Delta R)} f(\tilde{X}) d V(\tilde{X}) - \int_{\tilde{S}_{X} (R + \Delta R)} f(x) d V(\tilde{X}) - \int_{\tilde{S}_{X} (R)} f(\tilde{X}) d V (\tilde{X}) +  \int_{\tilde{S}_{X}(R)} f(x) d V(\tilde{X}) \right| \\
    & \leq & V_m C_{p, 2} ((R + \Delta R)^{m + 2} - R^{m + 2}) + V_m C_{p, 1} C_{Ric}((R + \Delta R)^{m + 3} - R^{m + 3}) \\
    & + & V_m C_{p, 2} C_{Ric} ((R + \Delta R)^{m + 4} - R^{m + 4}) \\
    & \leq & V_m \Delta R \left( R + \Delta R)^{m + 1} (C_{p, 2}(m + 2) + (m + 3) (R + \Delta R) C_{p,1} C_{Ric} + (m + 4)(R + \Delta R)^2 C_{p, 2} C_{Ric} \right).
  \end{EQA}

  Now, we find the error of the replacement of density with a constant in a small neighborhood of \(x\)
  \begin{EQA}
    & & \left| P(x \in \tilde{S}_{X} (R + \Delta R)) - P(x \in \tilde{S}_{X}(R)) - V_m m R^{m - 1} f(x) \right| \\
    & = & \left| \frac{\partial}{\partial R} \left( \int_{\tilde{S}_{X}(R)} f(x) d V(\tilde{X}) - \int_{A^{m - 1}} \int_0^R t^{m - 1} f(x) d t d \theta \right) \right|  \\
    & = & f \left| \int_{A^{m - 1}} \int_R^{R + \Delta R} t^{m - 1} f(x) d t d \theta \right| \leq f(x) V_m C_{Ric} ((R + \Delta R)^{m + 2} - R^{m + 2}) \\
    & \leq & f(x) V_m C_{Ric} \Delta R (R + \Delta R)^{m + 1}(m + 2).
  \end{EQA}

  By substituting all the obtained errors we find the estimator \(\lambda(R) = \frac{ \partial P(x \in S_x(R))}{\partial R}\mid_{R = t}\):
  \begin{EQA}
    & & \lim_{\Delta R \to 0} \frac{P(x \in S_{\tilde{X}} (R + \Delta R)) - P(x \in S_{\tilde{X}} (R))}{\Delta R} \\
    & = & V_m m R^{m - 1} f(x) + 8 V_m f_{max}(m + 2) R^{m + 1} \frac{m C_{II}}{24} + V_m R^{m + 1} (C_{p, 2}(m + 2) \\
    & + & (m + 3)R C_{p, 1} C_{Ric}(m + 4) R^2 C_{p, 2} C_{Ric}) + f(x) V_m C_{Ric} R^{m + 1} (m + 2) \\
    & = & R^{m - 1} V_m (m f(x) + R^2 \delta), 
  \end{EQA}
  where
  \begin{EQA}
    & & |\delta| \leq 8 f_{max} (m + 2) \frac{m C_{II}}{24} + C_{p, 2}(m + 2) + (m + 3) R C_{p, 1} C_{Ric} \\
    & + & (m + 4) R^2 C_{p, 2} C_{Ric} + f(x) C_{Ric} (m + 2).
  \end{EQA}

\subsection{Proof of Proposition~\ref{prop:geo_mle}}

  In order to consider geometric properties of a manifold, we replace the rate \(\lambda(R) = f(x) V_m m R^{m - 1}\) with the obtained estimate 
  \begin{EQA}[c]
    \lambda(R) = R^{m - 1} V_m (m f(x) + R^2 \delta):
  \end{EQA}
  in the log-likelihood function.
  \begin{EQA}
    & & L = \int_{0}^R \log \lambda(t) d N(t) - \int_0^R \lambda(t) d t  \\
    & = & (m - 1) \int_0^R \log t d N(t) + \log V_m \int_0^R d N(t) + \int_0^R \log(m f(x) + t^2 \delta) d N(t) \\
    & - & V_m m f(x) \int_0^R t^{m - 1} dt - V_m \delta \int_0^R t^{m + 1} d t = (m - 1) \int_0^R \log t d N(t) + N(R) \log V_m \\
    & + & \int_0^R \log(m f(x) - t^2 \delta) d N(t) - V_m R^m \left( f(x) + \frac{R^2 \delta}{m + 2} \right), 
  \end{EQA}
  which by Jensen's inequality is 
  \begin{EQA}
    & & L \geq \left\{\log(m f(x) + t^2 \delta) = \log 2 + \log \frac{m f(x) + t^2 \delta}{2} \geq \log 2 + \log m f(x) + \log t^2 \delta  \vphantom{\int_1^2} \right\} \\
    & \geq & (m - 1) \int_0^R \log t d N(t) + N(R) \log V_m + \log m \int_0^R d N(t) + \log m(x) \int_0^R d N(t) \\
    & + & \int_0^R \log 2+\log(t^2 \delta) d N(t) - V_m R^m \left( m(x) + \frac{R^2 \delta}{m + 2} \right) \\
    & = & (m - 1) \int_0^R \log t d N(t) + N(R) \log V_m
    + N(R) \log m + N(R) \log f(x) \\ & + & \int_0^R \log(2t^2 \delta) d N(t)- V_m R^m \left( f(x) + \frac{R^2 \delta}{m + 2} \right). 
  \end{EQA}
  We maximize the lower bound of the likelihood by \(\theta = \log f(x)\) and \(m\)
  \begin{EQA}
    & & \frac{\partial L}{\partial \theta} = N(R) - V_m R^m e^{\theta} \Rightarrow e^{\theta} = \frac{N(R)}{V_m R^m} \\
    & & \frac{\partial L}{\partial m}  = \int_0^R \log t d N(t) + \frac{V'_m}{V_m} N(R) + \frac{N(R)}{m} - V'_m R^m \frac{N(R)}{V_m R^m} \\
    & - & V_m R^m \frac{N(R)}{V_m R^m} \log R - \delta \frac{R^{m + 2} V_m}{(m + 2)^2} \left( \frac{V'_m (m + 2)}{V_m} + m + 2 - 1 \right) \\
    & = & \int_0^R \log t d N(t) + \frac{N(R)}{m} - N(R) \log R - \delta \frac{R^{m+2} V_m}{m + 2} \left( \frac{V'_m}{V_m} + 1 - \frac{1}{m + 2} \right).
  \end{EQA}
  \begin{EQA}
    & m = & \Bigg( \frac{1}{N(R,x)} \sum_{j=1}^{N(R,x)} \log \frac{R}{T_j(x)} \Bigg)^{-1} \\
    & \cdot & \left(1 + 
          2 \delta \frac{R^{m+2} V_m m}{N(R, x)(m + 2)} \left( \frac{V'_m}{V_m} + 1 - \frac{1}{m + 2} \right) \Bigg( \frac{1}{N(R,x)} \sum_{j=1}^{N(R,x)}\log \frac{R}{T_j(x)} \Bigg)\right).
  \end{EQA}
    
  Finally, we obtain for small \(R\)
  \begin{EQA}
    &&\hat{m}(x) = \left( 1 + \delta \frac{R^2}{N(R, x)} \right) \hat{m}_{\text{MLE}}(x).
  \end{EQA}
\end{document}